\documentclass{llncs}
\usepackage{amsmath} 
\usepackage{amssymb}
\usepackage{graphicx}
\usepackage{booktabs}
\usepackage{algorithm}

\usepackage{algpseudocode}% http://ctan.org/pkg/algorithmicx

\begin{document}
\def\code#1{\texttt{#1}}

\title{Dynamic Natural Language Processing with Recurrence Quantification Analysis}
\titlerunning{dNLP}  % abbreviated title (for running head)
\authorrunning{Dale \& Coco} % abbreviated author list (for running head)
\author{Rick Dale\inst{1}, Nicholas D. Duran \inst{2}, Moreno Coco\inst{3}}
\institute{Department of Communication\\University of California, Los Angeles \\
\email{rdale@ucla.edu}, \texttt{http://rdale.bol.ucla.edu}
\\
\and School of Social and Behavioral Sciences\\Arizona State University \\
\email{nduran4@asu.edu}, \texttt{http://dynamicog.org/} 
\\
\and Department of Psychology\\University of Edinburgh \\
\email{Moreno.Coco@ed.ac.uk}, \texttt{http://www.morenococo.org/}}

\maketitle              % typeset the title of the contribution

\begin{center}
\textbf{Draft, to be submitted. For full technical details,\\
including a working draft of the $\code{crqanlp}$ library,\\
please see $\code{http://github.com/racdale/crqanlp}$.}
\end{center}

\begin{abstract}
Writing and reading are dynamic processes. As an author composes a text, a \emph{sequence} of words is produced. This sequence is one that, the author hopes, causes a revisitation of certain thoughts and ideas in others. These processes of composition and revisitation by readers are ordered in time. This means that text itself can be investigated under the lens of dynamical systems. A common technique for analyzing the behavior of dynamical systems, known as recurrence quantification analysis (RQA), can be used as a method for analyzing sequential structure of text. RQA treats text as a sequential measurement, much like a time series, and can thus be seen as a kind of dynamic natural language processing (NLP). The extension has several benefits. Because it is part of a suite of time series analysis tools, many measures can be extracted in one common framework. Secondly, the measures have a close relationship with some commonly used measures from natural language processing. Finally, using recurrence analysis offers an opportunity expand analysis of text by developing theoretical descriptions derived from complex dynamic systems. We showcase an example analysis on 8,000 texts from the Gutenberg Project, compare it to well-known NLP approaches, and describe an R package (\code{crqanlp}) that can be used in conjunction with R library \code{crqa}.
\end{abstract}

\section{Introduction}
The distribution of words in a text has long been regarded as a statistical signature of cognition. These signatures relate to properties of writers \cite{crossley2011understanding} and readers \cite{mcnamara2007evaluating}, and can be used fruitfully in cognitive technologies \cite{graesser2011learning,mcnamara2013natural,roscoe2013writing}. Text also reflects a sequential structure that has cognitive or other psychological implications. In the space of applied natural language processing (NLP), a focus on the \emph{dynamics} of the text is much less common. From a text's words to its topics, a text reflects not only a writer's organization of thought, but presumably an attempt to generate similarly \emph{sequenced} thoughts in readers. For this reason, texts can also be treated as signatures of that dynamic cognitive process. This is the assumption we make here. We showcase how a method called recurrence quantification analysis (RQA) can be adapted to the analysis of text. Measures that RQA provides have direct parallels in traditional natural language processing, and we derive these equivalences below. Importantly, we argue that RQA offers a kind of conceptual framework in which to think about text as a dynamic process. 

In what follows, we briefly review a number of other dynamic approaches to text. These approaches together with RQA may be grouped into a general class of dynamic natural language processing techniques. We then briefly introduce relevant areas of applied natural language processing that will be used as a comparison framework, namely classic $n$-gram models of language. We then summarize RQA and showcase how it produces measures that are related to, and expand, measures from traditional NLP. We provide a detailed study of an example analysis, showing that this dynamic interpretation of written text can classify 8,000 samples from the Gutenberg Project.  

\subsection{Text as dynamics}

\noindent Several examples of recent research serve to illustrate ways that text can be treated as a kind of cognitive dynamics. For example, Doxas and colleagues \cite{doxas2010} use semantic modeling and analyze how texts traverse a semantic space, word by word. They find a law of scaling across a text that is consistent across many languages, and describe it as ``dimensionality of discourse.'' Altmann and colleagues have recently extensively studied the manner in which content vs. function words occur across a text, and used statistical mechanics from physics to describe text properties \cite{altmann2009beyond,altmann2012origin}. They have shown how words exhibit ``bursty'' regularities, in how they recur in texts. In a similar vein, the rich text-permutation analyses of Moscoso del Prado Mart\'{i}n \cite{del2011universal} show how text reveals levels of dynamic control, in accord with levels of linguistic analysis: from orthographic units to discourse. These dynamic concepts have also been extended to analysis of transcripts \cite{angus2013making,butner2008facts,dale2005categorical,de2014visualising}

The analysis we show here is an instance of this ``text as dynamics'' approach. The first study to use RQA on text is Orsucci and colleagues \cite{orsucci1997orthographic}, in which they show that certain cultural patterns can be revealed in the dynamic patterns of poetry. Dale and Spivey \cite{dale2005categorical,dale2006unraveling} develop a general framework for applying RQA to transcripts and corpora, and show that dynamic analysis of transcripts may reveal language development and social alignment. A number of other projects have utilized RQA on text as well. Wallot \cite{wallot2017recurrence} shares an elegant tutorial for using RQA in discourse data, and Allen and colleagues have recently used RQA to analyze writing samples from students \cite{allen2017recurrence}.

The purpose of the present paper is to situate RQA in broader NLP context, and offer a set of conceptual and computational tools for doing this kind of dynamic natural language processing. We show that a common traditional NLP analysis, the $n$-gram framework, has particular equivalences to RQA. This bridge helps to understand RQA and how it works. In addition, it will show that RQA provides subtly different measures to quantify the dynamics of text, and also provides a conceptual framework, namely dynamical systems, for thinking about text in a different way. Next, we summarize the basic structure of $n$-gram models to set the stage for this comparison. 

\subsection{$n$-gram models}

A classic way of modeling natural language is the probability distribution over sequences of length $n$ of that language \cite{shannon1951prediction}. Such a model, despite its simplicity, can be used for text generation, guides for speech recognition systems, powerful baseline models for linguistics, and, classically, as a framework for understanding information theory \cite[for review]{jurafskymartin,manning1999foundations}. We assume the reader has some familiarity with $n$-gram models, but we recapitulate some key features that will be useful as a comparison.

For our purposes, we take \emph{words} to be the units over which we compute probabilities (though any symbol sequences can be used). A model is the probability distribution of strings of length $n$ from a corpus. A bigram model, for example, takes $n=2$ and is the probability distribution over words conditioned on a single prior word context, defined over some text $S$. In general terms:

%https://oeis.org/wiki/List_of_LaTeX_mathematical_symbols#Set_and.2For_logic_notation

\begin{equation} 
	ng_k(S) = \{P(w_i|C(k-1)) : \forall w_i \in S\}
\end{equation}

\noindent Here $C(k-1)$ denotes some context of length $k-1$. This is commonly taken to be some sequence of prior words leading to the $i^{th}$ word $w_i$: 

\begin{equation} 
P(w_i|w_{i-k+1},...,w_{i-1}) = 
	\frac{P(w_i,w_{i-k+1},...,w_{i-1})}{P(w_{i-k+1},...,w_{i-1})}
\end{equation}

\noindent As it will be useful later, this can also be expressed in terms of the frequency of the $n$-grams in the given corpus:

\begin{equation}
P(w_i|w_{i-k+1},...,w_{i-1}) = 
	\frac{f(w_i,w_{i-k+1},...,w_{i-1})}{f(w_{i-k+1},...,w_{i-1})}
\end{equation}

\noindent Considerable work on $n$-gram models emerges from both theoretical and practical application of them. For example, we omit various obvious issues that can arise, such as unencountered words when using $ng_k$ in various ways, and the reader can consult the excellent introduction in \cite{jurafskymartin} for summary of these issues. Of course, most of these probabilities are computed using relative frequency. In the bigram case, each entry of $ng_2(S)$ is $f(w_{i-1},w_i)/f(w_{i-1})$, where $f(x,y)$ is the number of times the string ``$x$ $y$'' occurs in $S$. To improve model performance, there are a number of techniques for ``smoothing'' the distribution so that estimations of even unseen combinations can earn a bit of the probability mass, the most common among them the Kneser-Ney method \cite{chen1996empirical,kneser1995improved}.  

Despite the model's simplicity, much can be done with it. It is important to distinguish the set of estimated probabilities from algorithms developed to capitalize upon them. For example, text-generation schemes can use $ng_k(S)$ to sample a sequence of words guided by these probabilities. Doing so generally reveals that higher-order $n$-gram models perform more compellingly than lower-order ones, though they become more fixed to the training corpus. $ng_k$, even when just $k=1$ or $2$, can be used to estimate the probability of sentences unseen in a training corpus, which can be useful for a variety of reasons, such as estimating linguistic judgments \cite{reali2005uncovering}. 

Though $n$-gram models permit assessment of specific individual sequences of words, here we use these models for  aggregate description of texts. This will permit direct comparison to measures obtained from the dynamic method, RQA. We could take, for example, first-order Shannon entropy of a given text relative to $ng_k$ as the expected value of the self-information of each $k$-gram, $1/|ng_k|\sum^{P \in ng_k} P \mbox{log} P$. We will revisit measures such as this below, and provide detail where appropriate.

\section{Recurrence Quantification Analysis}

A relatively new time series analysis, called recurrence quantification analysis (RQA), has now been used to study many different physical \cite{marwan2002nonlinear} and biological \cite{webber1994dynamical} systems, including to model mathematical systems \cite{trulla1996recurrence}. The method has been described as a kind of generalized cross-correlation analysis \cite{marwan2007recurrence}, and it provides new descriptive measures to summarize a time series \cite[for a review and summary]{marwan2008historical}. These measures include the relative deterministic properties of the time series, how much drift is present, and so on. RQA has been shown to describe properties that linear methods, such as correlation-based techniques, cannot always capture \cite[Fig. 20]{marwan2007recurrence}.

RQA quickly gained traction in the analysis of time series in cognitive science, such as in the domain of motor control \cite{riley1999recurrence,shockley2003mutual}. It was first used for text analysis by Orsucci and colleagues \cite{orsucci1997orthographic} in an analysis of poems. Orsucci et al. describe how the time series technique can be adapted for the analysis of character sequences, but any categorical (``nominal'') behavior sequences can be used in conjunction with RQA \cite[for generalizations and descriptions]{dale2005categorical,dale2011nominal,wallot2017recurrence} see REF; REF; Seb). Rather than explain RQA in the more common domain of continuous, physical time series \cite[for summary]{webber2005recurrence}, we will immediately explain how it works on text. This serves our purpose here for obvious reasons. However, understanding how RQA works on continuous time series is a relatively simple extension of the categorical case. To explain RQA, we will describe it as an analysis of two steps: (i) building a recurrence plot (RP), then (ii) quantification recurrence on that plot (RQA).

\subsection{Recurrence Plot (RP)}

The recurrence plot is simple to describe in formal terms. Consider a sequence of categories or codes $A$ of length $N$, $x(t) \in A$. These may be characters \cite{orsucci1997orthographic}, words \cite{dale2005categorical}, syntactic classes \cite{dale2006unraveling}, emotion categories \cite{main2016exploratory}, or regions of interest from an eye tracker \cite{richardson2005looking}. A \emph{recurrence} is defined as a repetition of a given code at two points in time $i$ and $j$. The RP is simply a visualization on the plane of all such repetitions or recurrence $(i,j)$: $RP = \{(i,j) : x(i) = x(j)\}$. This can be generalized further, as described in \cite{dale2005categorical}. An RP can be seen as a set of points $(i,j)$ of a time series such that some function $F$ over that time series satisfies some relation $R$: 

\begin{equation}
	RP = \{(i,j) : R(F(x(i)),F(x(j))) \rightarrow \top \}
\end{equation}

\noindent In our case, we will take codes reflecting words $w_t$, the sequence to be a text $S$, a relation $R$ of equivalence ($=$), and set the function $F$ simply to the identity function for the word sequence $w_i$, so that $F(w_i) = w_i$ and we obtain the simplest definition of a $RP$ $\{(i,j) : w_i=w_j \}$. As we briefly describe below, variants of RPs are possible by modifying these definitions, and many have been proposed \cite{marwan2007recurrence}. The general definition will allow us to revisit the case of continuous data later in this paper (where our sequence consists of scalar observations, rather than codes, such as words). With this simple definition of a recurrence plot over text, let us visualize a sample text. One of the texts used in our example analysis below comes from the Gutenberg Project, the 1906 book \emph{A Kindergarten Story} by Jane Hoxie (Gutenberg Project ID: 14127). The plot clearly shows dense regions of word reuse from story segment to segment, with long stretches of diagonal lines. In fact, this text for children has the highest $DET$ score of the thousands of texts we analyze below (see genre analysis below for details on how we pre-processed this text).

\begin{figure}
  \centering
    \includegraphics[width=0.5\textwidth]{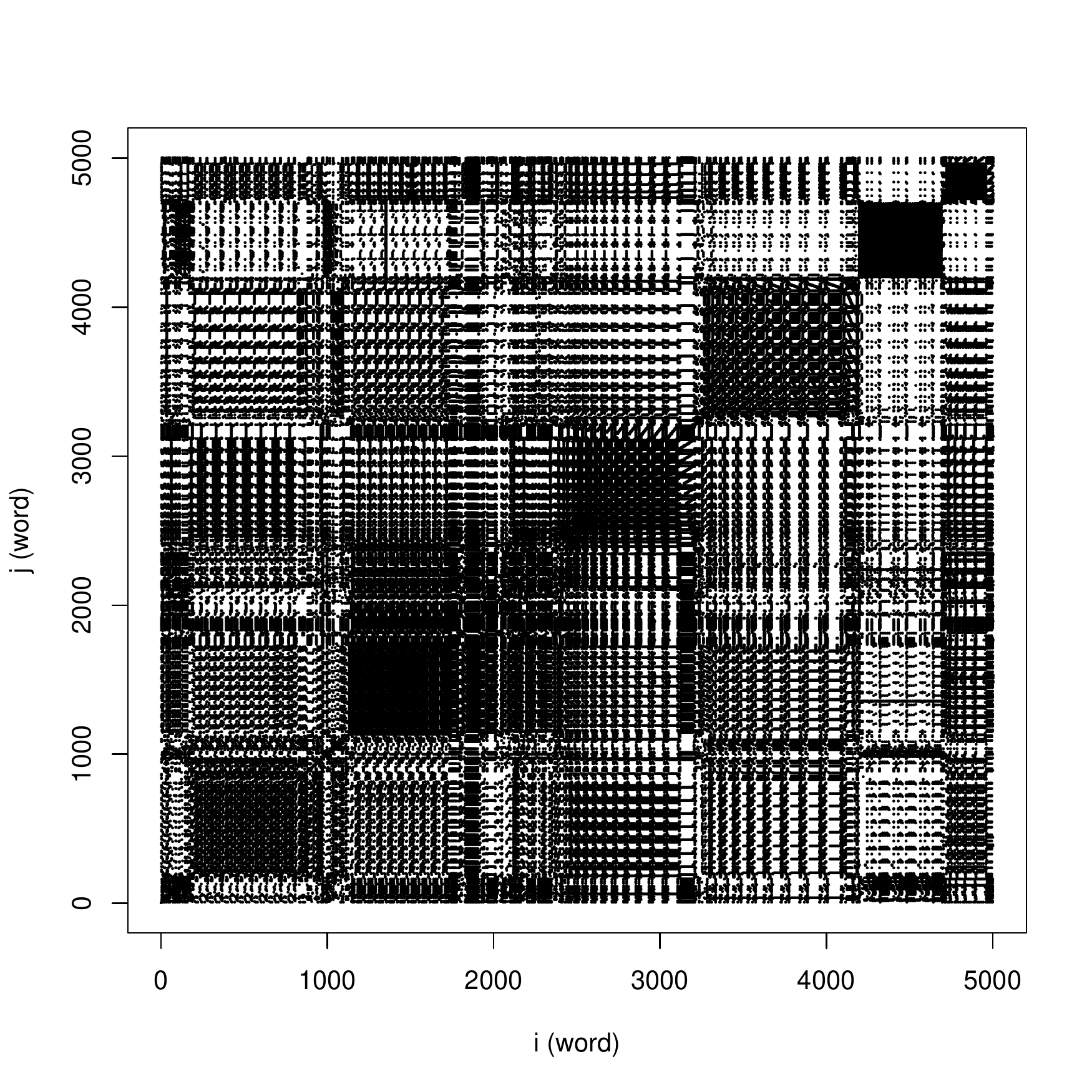}
  \caption{Sample $RP$ of the text \emph{A Kindergarten Story} by Jane Hoxie.}    
  \label{sample_rp}
\end{figure}

\noindent This visualization will be familiar to many, as it is equivalent to Church dot-plot methods \cite{church1993dotplot}. In the emerging tradition of recurrence quantification, a specific suite of measures is extracted from this plot. These measures provide new descriptions, and have a dynamic interpretation -- how the time series (in our case, text) is changing in time.

\subsection{Recurrence Quantification (RQA)}

Once an $RP$ is defined, RQA involves quantification of the plot: the number of points, and the nature of their distribution. This quantification typically comes in the form of a suite of measures that describe the $RP$. Several measures are correlated, but they can be interpreted subtly differently. We introduce the most common measures taken from the plot here. These will serve as our anchor to the $n$-gram analyses, showing patterns of equivalence to traditional NLP.

\subsubsection{Recurrence Rate ($RR$)} Recurrence rate ($RR$) is the proportion of points on the plot. It is the simplest measure, computed by taking the number of points $|RP|$ and divide this by the number of possible points $N^2$. It can be useful to consider $RR$ to be the result of a nested sum of this sort, spanning rows and columns of the $RP$:

\begin{equation}
RR = \frac{\sum_{i=1}^N \sum_{j \neq i}^N m_{RP}(i,j)}{N^2}
\label{eqn:RRRQA}
\end{equation}

\noindent Where $m_{RP}$ is a membership function for a given $RP$, returning the value $1$ if membership is true, and $0$ otherwise. Note that the cases of $i=j$ are often excluded from this calculation because recurrence is trivial when we compare a series to itself at the same time indices, known as the ``Theiler window'' \cite{hegger1999practical}.

\subsubsection{Determinism ($DET$)} 
\begin{figure}
  \centering
    \includegraphics[width=1\textwidth]{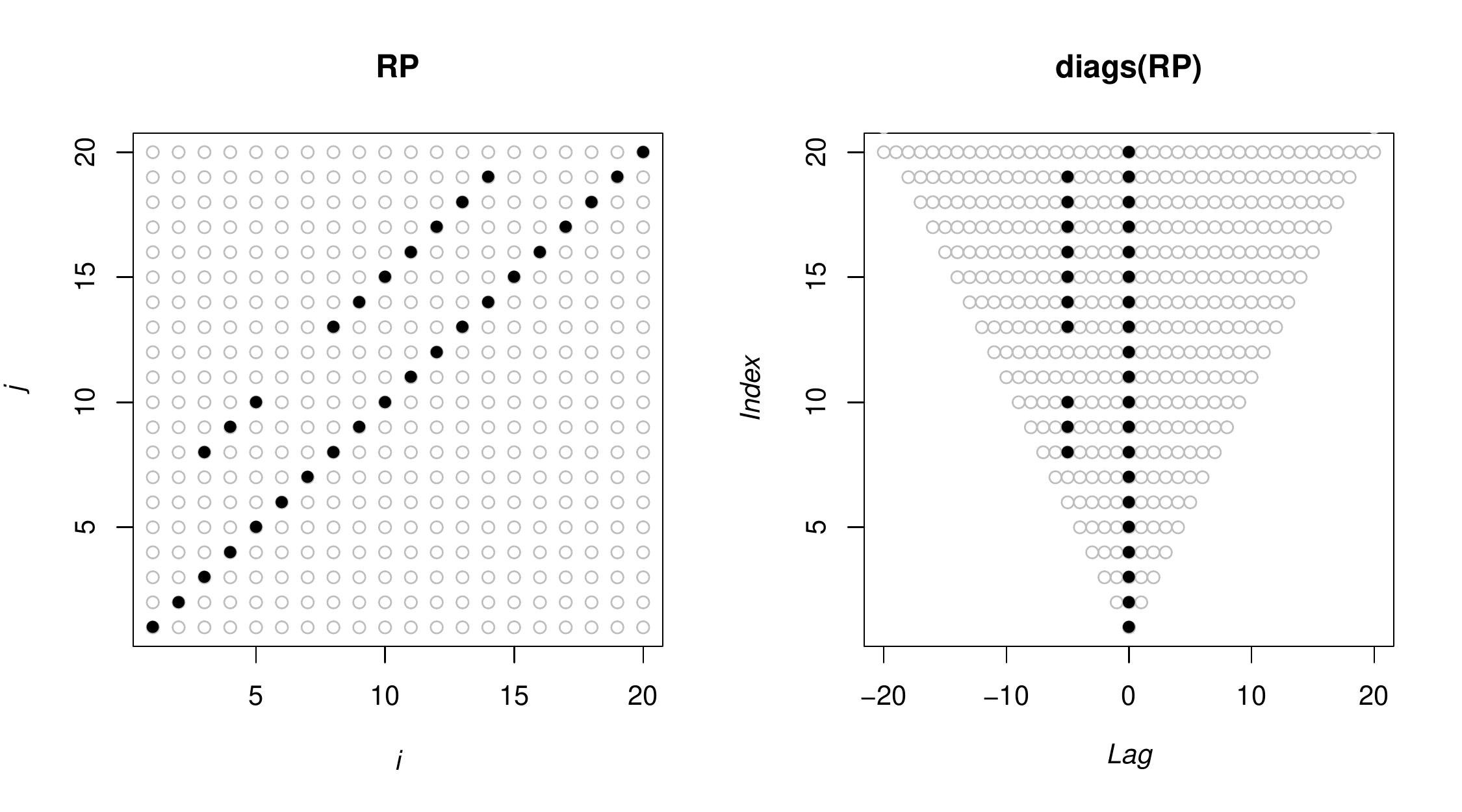}
  \caption{Illustration of computing $DET$ by rotating along diagonals.}    
  \label{det_rotation}
\end{figure}

$RR$ does not reveal \emph{how} these points are distributed. In the first construction of the RP visualization, Eckmann and colleagues \cite{eckmann1987recurrence} noted that points tend to distribute in interesting ``textures.'' Determinism ($DET$) captures one such texture: to what extent points line up on diagonal lines. Diagonal lines reflect \emph{paths} that are being revisited by the time series. This is shown above in the example $RP$, in which a particular sequence of words in the Iliad is being repeated quite often (this tends to be reference to Achilles and his estimable pedigree and such). Obtaining these lines can be easily defined algorithmically. This definition will be useful for showing equivalence to measures based on $n$-gram analysis. To compute $DET$, we can represent the diagonals on the $RP$ as columns of a new matrix, $\code{diags}(RP)$. This is illustrated in Fig. \ref{det_rotation}.

Using these columns of the rotated $RP$, $DET$ can now be computed as the proportion of points that fall inside columnar runs of length 2 or more. To identify runs, we can extract the set of all column indices at which points are \emph{not} present (i.e., 0). As illustration, define the function $\code{i}(Lag(k))$ as one that extracts these indices at which 0s can be found along the column (lag) $k$: $\code{i}(Lag(k)) = \{Index : \code{diags}(RP)_{k,Index} \notin RP\}$. We then difference these ordered indices, with the obvious outcome that $\Delta\code{i}(Lag(k)) \geq 1$. When $\Delta\code{i}(Lag)=1$, there are adjacent 0s (absence of points) along the column. We therefore want to obtain intervals between these cases of absence, reflecting points in time at which the 0 indices ``jump'' over the runs of 1 (reflecting, instead, adjacent points). The length of a diagonal line is the magnitude of this difference, and $DET$ can be computed simply from this distribution of line lengths. Consider the example from Fig. 2, left. In this plot, the lag of -5 contains 2 line segments:
 
\begin{align*}
Lag(-5): 0\overbrace{111111}^\text{line 1} 
00\underbrace{111}_\text{line 2}00
\\
\code{i}(Lag(-5)) = 
<1,8,9,13,14>
\\
\Delta\code{i}(Lag(-5)) = <7,1,4,1>
\end{align*}

\noindent The set of line segments is thus the set $ lines = \{\Delta\code{i}(Lag(k))-1 : \Delta\code{i}(Lag(k))>2,k \in [-N:N]\}$. A critical ingredient of this analysis of the $RP$ is that it \emph{uniquely} delimits line lengths. Each line is identified as a single continual sequence, and potential subsequences are bounded therein. Even though, by definition of a line segment, a sequence of 10 points contains 9 points of length 2, the computation segments \emph{all} adjacent points together. This will help us compare to $n$-gram models in an elegant way. Once this set of lines is obtained, $DET$ is computed as:

\begin{equation}
DET = |RP|^{-1}\sum_{i=1}^{|lines|}lines_i
\end{equation}

\subsubsection{Entropy ($ENT$)} An additional measure we will define is the entropy of the diagonal line segments. This was introduced to describe the manner in which a system may be revisiting particular sequences. If a system is highly regular, diagonal lines will distribute over a characteristic limited number of line lengths. In systems that are more disordered or variable in some way, the distribution over line lengths will be more uniform. In RQA, we simply take the Shannon entropy of the $lines$ set defined above. Take $p(l)$ to be the proportion of lines of length $l$. $ENT$ is defined as $-\sum_{l=1}^{\code{max}(lines)} p(l)log(p(l))$.

\subsubsection{Other measures computed from $DET$} We summarize a few other measures, because they are a simple extension of $DET$. A common measure used in RQA is the maximum line segment ($MAXLINE$), defined simply as $\code{max}(lines)$, and mean line ($MEANLINE$) as $|lines|^{-1}\sum lines$. Some have also fruitfully quantified the vertical patterns on the plot, but we omit consideration of these here.

An interpretation of all these measures is described in Table 1. We omit a number of other commonly used measures for the sake of simplicity. These are summarized in \cite{coco2014cross}.

\renewcommand{\arraystretch}{2.0}
\begin{table}[htbp]
\centering\scriptsize
\caption{Summary of RQA measures considered here}
\label{my-label}
%\begin{tabular}{@{}lll@{}}
\begin{tabular}{@{}lp{4cm}p{4cm}@{}}
\toprule
\textbf{RQA Measure} & \textbf{Summary}                                                                     & \textbf{Sample Interpretation}                                                       \\ \midrule
Recurrence Rate (RR) & The percentage of time that the text revisits the same unit of analysis (e.g., word) & How much does an author re-use the same words?                              \\
Determinism (DET)    & The percentage the recurrent states (e.g., words) fall on paths of length 2 or more  & How much do words tend to fall on repeated sequences of the same phrase?    \\
Entropy (ENT)        & The Shannon entropy of the line length distribution                                  & How ordered (or disordered) are the repeated sequences?                     \\
Maximum line length (MAXLINE)  & Maximum line length in $lines$ set                                                   & What is the longest sequence of words that the author repeats, verbatim?    \\
Mean line length (MEANLINE)     & Mean value of the $lines$ set                                                        & What is the average length, in words, of a sequence that an author repeats? \\ \bottomrule
\end{tabular}
\end{table}

\subsection{Relevance to cognitive dynamics}

As we show below, these measures from RQA have equivalence relations to measures
extracted from $n$-gram models. But they are different from traditional models of this sort. They offer a distinct ``dynamic lens,''
through which we can interpret text sequences. This lens is relevant to the
composition of text, and the intended readers of that text. The intended audience for a piece of writing may influence its writer by constraining the dynamics that writer wishes to establish in her audience. Writing for children constrains word choice, sentence structure, and so on; writing modern poetry, on the other hand, may involve different sorts of dynamic wordplay. 
These features of a text can be measured using $n$-gram and related models -- but
we can also express findings in terms of our dynamic lens, through recurrent patterns quantifiable in text. There are a few benefits
of this lens that we consider here before moving forward.

\begin{itemize}
  \item Analyzing text with RQA links analysis to the broad literature in which RQA is used. Just as using statistical mechanics to analyze text can help bridge cognitive systems to physical systems \cite{altmann2009beyond}, using RQA to analyze text, and summarizing results in terms of its measures, allows us to compare effects to a very broad range of systems studied using this technique \cite{marwan2008historical,marwan2007recurrence}
  \item The measures from RQA are not transparently translatable into $n$-gram models. They do have  features that make them a unique. For example, as we show below, $DET$ measures bounded recurrent trajectories, and $ENT$ measures the order/disorder of these bounded trajectories.
  \item The growing demand for multimodal analysis invites frameworks that can integrate quite different kinds of variables. RQA is capable of working on any kind of time series or data sequence with few statistical assumptions, including verbal and non-verbal modalities that have differing temporal properties \cite{louwerse2012behavior}.
\end{itemize}

\section{Categorical RQA $\subset$ $n$-grams}

RQA measures can be extracted from $n$-gram models. Here we show the equivalence of three main measures: $RR$, $DET$, and $ENT$. It is important to note that the equivalence relation here is for \emph{categorical} RQA, namely analysis of a fixed discrete set of units, such as orthography or lexical items. This relationship does not hold for continuous data when subjected to RQA, but a comparison will still be interesting. We consider this later in the paper, but begin with a demonstration that RQA $\subset$ $n$-grams for basic text analysis.

\subsection{$RR$ and unigram frequency}

The simplest $n$-gram model over string $S$, $ng_1(S)$, is known as a unigram model, and is the set $ng_1(S) = \{P(w_i) : \forall w_i \in S\}$. To recover the counts of each word, we multiply the member probabilities of this set. We will represent this as $N ng_1(S)$. Recall from the definition above that $RR$ is computed with numerator $\sum_i\sum_{i \neq j}m_{RP}(i,j)$. For a given word, it will be counted in the $RP$ as many times as it occurs in $S$, minus 1, because we discount the recurrence point that reflects identity. Because we take the sum over \emph{all} word types in S, we have the product $N ng_1(S)(N ng_1(S)-1)$, implied also by the double sum. This recovers the numerator. With simple rearrangement we have the following equivalence with $RR$:

\begin{equation}
RR = N^{-1}ng_1(S)(Nng_1(S)-1)
\end{equation}

\noindent Or more intuitively:

\begin{equation}
RR = N^{-2}\sum_i^{w_i \in S}f(w_i)(f(w_i)-1)
\label{eqn:RRintuitive}
\end{equation}

\noindent This means that $RR$ is a scaled Euclidean norm of the observed word frequencies. If we assume uniform probability of words making up $S$, then $RR$ drops as a function of the square of the type-token ratio. If we take the cardinality $B = |ng_1(S)|$ as the number of types, then assuming uniformity we have $RR \approx 1/|ng_1(S)|$:

%\begin{align}
%Sp(k) &= \sum^{NG_k(S)} [(N-k+1)P(w_t|w_{t-1}...w_{t-k+1})-1] \\
%&= \sum^{NG_k(S)} [\frac{C(w_t,...,w_{t-k+1})}{P(w_{t-1},...,w_{t-k+1})}-1]
%\end{align}

\begin{align*}
RR &= N^{-2}\sum_i^{w_i \in S}f(w_i)(f(w_i)-1) \\
&= N^{-2} B \Big[\frac{N}{B}\Big(\frac{N}{B}-1\Big)\Big], \mbox{due to uniformity} \\
&= \frac{1}{B} - \frac{1}{N}
\end{align*}

\noindent Under assumptions of uniformity, $RR$ reduces to a simple function of the cardinality of the $n$-gram model -- the type-token ratio. This does not hold when uniformity is violated, of course. With $p_i$ representing the proportion of word $i$ to simplify notation, two word types generates the following $RR$ in a text: 

\begin{align*}
RR &= N^{-2}\Big[Np_1(Np_1-1)+Np_2(Np_2-1)\Big]\\
&= N^{-1}\Big[Np_1^2-p_1+Np_2^2-p_2\Big]\\
&= p_1^2+p_2^2-\frac{1}{N}
\end{align*}

\noindent Again reflecting its status as a kind of Euclidean norm. The prior two formulations suggest, and it is easy to show, that in general $RR = \sum p_i^2-1/N$. In any case, it is evident that $RR$ can be computed as a property of the unigram model for string $S$, $ng_1(S)$.

\subsection{$DET$ and maximally bounding $n$-grams}

$RR$ can be easily derived from unigram frequencies, and seen as a kind of Euclidean norm capturing the tendency for a text's elements to recur in time. $DET$ is more complicated, because it captures the sequences underlying the text. We could imagine $DET$ as a measure that quantifies tendencies for the text to organize itself into sequences. One intuitive expectation is that $DET \approx |RP|^{-1}\sum_{k=1}^{\code{max}(lines)}ng_k(S)$, but this is not the case. The relationship is not exact, because recall from the definition of $DET$ that it defines lines in a manner that \emph{maximally bounds} other candidate sequences. For this reason, sequences of length $k-1$, if contained in longer sequences of length $k$ are not included separately in the tally. $DET$ is therefore a statistic reflecting the extent to which recurrent words (or letters, etc.) fall on maximally bounding $n$-grams. We can demonstrate this algorithmically, using the strategy shown in algorithm 1. 

\begin{algorithm}
\caption{Counting maximally bounding $n$-grams}
\begin{algorithmic}[1]
  \For{$k$ in $N-1$ to $2$}
  	\State Identify set of unique $k$-grams
    \State Set $DET_k=0$
    \For{all $k$-grams}
    	\State $f$ = Count $k$-gram
        \State Discount $f$ by $|$grep($k$-gram,$j$-grams where $j>k$)$|$
        \State $DET_k= DET_k + f^2-f$
    \EndFor
    \State Store $(k,DET_k)$ pair
  \EndFor
\end{algorithmic}
\label{algoDET}
\end{algorithm}

\noindent Note that line 6 is the critical line which ensures that a higher-order $j$-gram is maximally bounding -- none of its constituent $k$-grams are counted. However, there may still be constituent $k$-grams occurring outside of that bounding context. This can be seen as a discounting factor that ensures no double counts. On line 7, we use $f^2-f$ under the same logic as the double sum in Eq. \eqref{eqn:RRRQA} above. 

$DET$ is now simply the outer product of line lengths and the stored (ordered) discounted frequencies $DET_k$, divided by the total number of unigram points, implied by Eq. \eqref{eqn:RRintuitive} above:

\begin{equation}
DET = \frac{\sum_{k=2}^{N-1}k DET_k}{
	\sum_i^{w_i \in S}f(w_i)(f(w_i)-1)
    }
\end{equation}

\subsection{Shannon entropy of maximally bounding $n$-grams}

We require the distribution stored in the $(k,DET_k)$ pairs because calculating $ENT$ uses this distribution. It is the Shannon entropy over the distribution of maximally-bounding $n$-gram lengths. It is therefore a measure of the disorder of the strings that maximally bound any other repeating $n$-grams in $S$. It can be simply calculated by normalization of the $DET_k$ values: 

\begin{equation}
ENT = -\sum^{DET_k>0}\frac{DET_k}{N_l} \code{log}\frac{DET_k}{N_l}
\end{equation}

\noindent Where $N_l$ is equal to the total number of lines inferred by Algorithm \ref{algoDET} above.

\section{Interim summary and implications}

Categorical recurrence quantification, using discrete categories as the basis for the plot $RP$ and its quantification via $RR$, $DET$, etc., is entirely reducible to information in traditional $n$-gram models. The $2D$ expansion of events onto the $RP$, out of string $S$, produces a kind of Euclidean norm over frequencies. $DET$ is the percentage of tokens that fall on maximally bounding $n$-grams. $ENT$ is the Shannon entropy of the probability distribution of maximally bounding $n$-gram lengths. To those familiar with dot-plot matrices and similar representations, this may come as no surprise \cite{church1993dotplot}. To those who have used recurrence with text, it may also be an intuitive relationship. The foregoing demonstrations show \emph{how} the measures traditionally used in RQA relate to frequencies of sequences, in a string's $n$-grams.

Despite this equivalence, RQA offers measures that are not common in $n$-gram models, and provide information that may supplement these traditional measures. $RR$ is not simply a linear function of token frequencies in $S$. It is a quadratic transformation, encoding deviation from uniformity. $RR$ is therefore related to other statistics that relate deviation from uniformity -- such as the $\chi^2$ statistic. Using the observed count formulation for chi-square, consider the following, where the observed count for any word $w_i$ is $f_i$:

\begin{align*}
\chi^2 &= \sum^B_i \frac{(f_i-N/B)^2}{N/B} \\
%&= \frac{B}{N}\Big(\sum^B_i f_i^2 + 
        %\sum^B_i \frac{N^2}{B^2} - 2\frac{N}{B} \sum^B_i f_i \Big) \\
&= \frac{B}{N}\Big(\sum^B_i f_i^2 + 
        \frac{N^2}{B} - 2\frac{N^2}{B} \Big) \\
&= N \cdot B\Big(\sum^B_i f_i^2/N^2 - 
        \frac{1}{B} \Big)  \\
&= N \cdot B\Big(\sum^B_i p^2 - 
        \frac{1}{B} \Big)    \\
&= N \cdot B \sum^B_i p^2 - N     
= N \cdot B \cdot RR + B - N
\end{align*}

\noindent This equivalence demonstrates a statistical link to the simplest $RQA$ measure, $RR$, as estimating deviation from uniformity. It is debatable whether such an equivalence will render a new measure based on this intuition alone because, of course, texts will strongly deviate from uniformity in all reasonable cases. However the extent to which this deviation occurs may covary interestingly with features of a text, such as genre or difficulty level. We consider this below. 

\begin{figure}
\vspace*{-0.5in}
  \centering
    \includegraphics[width=0.6\textwidth]{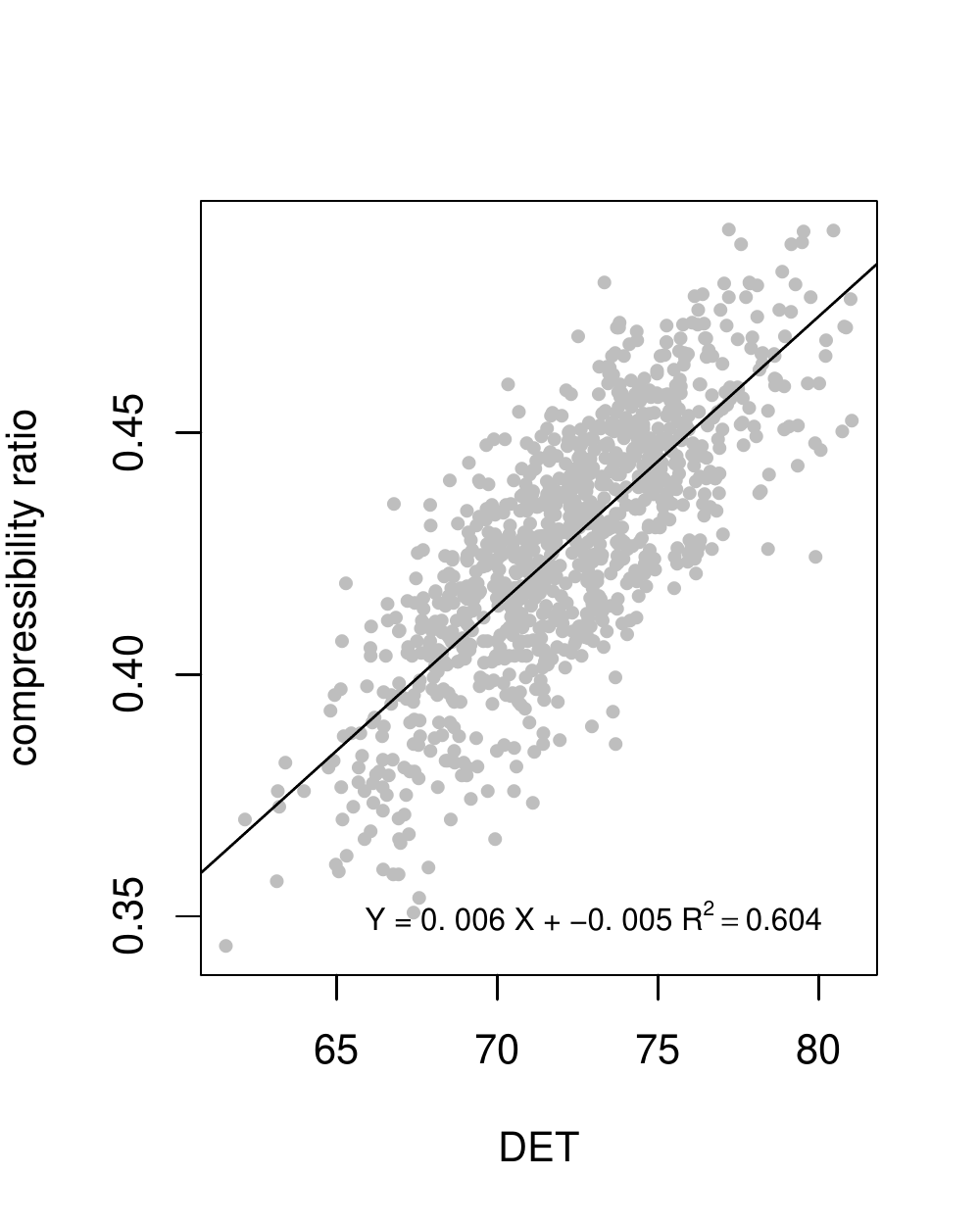}
  \caption{Relationship between $DET$ and compression ratio. $S$ was generated by randomly sampling 100 times from substrings 'a','b','c','a b','b c', and 'a b c'. The figure is based on 1,000 such generations of random $S$.}    
  \label{fg:compression}
\end{figure}

$DET$ and $ENT$ are based on maximally bounding $n$-grams. These are bounded paths that most efficiently reflect the recurrent paths in the text.  Note however that they do not include a conditionalization of the sequences themselves -- only the count statistics of the $n$-grams. Measures deriving from $DET$ have natural interpretation too. $MAXLINE$ is the easiest: It is the longest subsequence that occurs more than once. There is an additional implication of $DET$ as based on maximally bounded $n$-grams. By seeking out the most efficient means of encoding \emph{repeated sequences}, $DET$ correlates with compressibility ratio. This is shown in Fig. \ref{fg:compression} on randomly generated strings $S$. $DET$ describes the ``compressibility'' of the system's behavior.

\begin{equation*}
\mbox{compressibility ratio} = 1-\frac{\code{Size}(\code{Zip}(S))}{\code{Size}(S)}
\end{equation*}

In the next section, we briefly describe extensions of RQA and how they relate to text. These are also reducible to $n$-gram models but they further highlight that $RQA$ is a framework for treating text as a kind of dynamics.

\section{Extensions: $TREND$, Windowed RQA, and CRQA}

We have introduced primary measures commonly used in application of RQA to various time signals: $RR, DET, EMT, MEANLINE$, and $MAXLINE$. There are many extensions of RQA, and these extensions go beyond any obvious equivalence to traditional statistical models \cite{marwan2007recurrence}, including $n$-gram models.

First, $TREND$ is traditionally interpreted as the amount of drift that a time series exhibits. It is measured by looking at how the $RP$ ``whitens'' as one moves perpendicularly from the line of identity, where $i=j$. Put differently, it is the tendency for recurrence to drop as $|i-j|$ increases, and is computed using a regression slope. Adapted from \cite{marwan2007recurrence}, it is formulated as:

\begin{equation*}
TREND = \frac{\sum_{i=1}^{\tilde N}(i-\tilde{N}/2)(RR_i-E[RR_i])}
			{\sum_{i=1}^{\tilde N}(i-\tilde{N}/2)^2}
\end{equation*}

\noindent Here, $RR_i$ is the percentage recurrence along a diagonal $i$ relative to the line of identity, where $Lag$, or $i$, is $0$. When $TREND<0$, it reflects a system that has diminishing recurrence as time indices separate, meaning that $RR_i$ is dropping. In text, this may reflect the number of topics in a text and how quickly the text transitions through them.

Windowed RQA is an extension of the standard approach by taking segments of a time series (or text) and performing RQA on each section. It has been shown that RQA measures, as they change across a time series, may mark transitions in the dynamic behavior of a system. For example, by studying the famous logistic map, Trulla and colleagues \cite{trulla1996recurrence} show that RQA is highly sensitive to bifurcation and onsets to chaos. Running windowed RQA on text is rather straightforward, and may also reflect changes in stylistic or topical regime in the text. We show an example of this in the next section, showing how segmentation of a text, and applying RQA measures to each, may shed interesting light on how a text is changing.

Finally, RQA may be used with two time series, and this method is referred to as cross recurrence quantification analysis (CRQA). In this case, a \emph{cross} recurrence plot is defined as the set points $(i,j)$ such that the states from two separate time series are recurrent. CRQA has been shown as a kind of generalized lag sequential method \cite{dale2011nominal}. We do not consider this extension in detail here, but note that this rather simple modification of RQA permits the exploration of conversational processes \cite{dale2005categorical,dale2011nominal} and in the case of NLP, analyses that may relate to document alignment procedures in other domains \cite{kay1993text}, and sequence alignment in bioinformatics \cite{kumar2004mega3}.

\section{Sample Text with \code{crqanlp}}

The library $\code{crqanlp}$ contains a number of functions, wrapped around library $\code{crqa}$, to facilitate rapid analysis of text. To illustrate this, we show a few lines of code processing the text from Fig. \ref{sample_rp}, the text \emph{A Kindergarten Story} by Jane Hoxie. This shows clustered regions of recurrence -- subsidiary tales in this children's book from 1906 (Gutenberg ID = 14127). The following shows how the library loads in the text, generates an analysis, plots the $RP$, and conducts windowed recurrence. Further options and summary of the code can be found on the GitHub repository: $\code{https://github.com/racdale/crqanlp}$.

\begin{verbatim}
   a = text_rqa('14127.txt',typ='file',embed=1,tw=0)
   plot_rp(a$RP,cex=.25,xlab='i (word)',ylab='j (word)')
   a = text_win_rqa('14127.txt',typ='file',winsz=500,wshft=20)
\end{verbatim}

\section{Sample Application of RQA: Genre}

Using the library \code{gutenbergr}, we extracted 6,408 texts from the most frequent 20 subjects. These are listed in Table \ref{tb:rqameasures} \cite{hlavac2013stargazer}. We restricted texts to those that had all measures defined (e.g., at least one diagonal line), could be downloaded in under 5 seconds from the library, and contained at least 10,000 words\footnote{In some cases, text formatting issues keep some \code{gutenbergr} functions from completing, in particular the ``strip'' function, so we constrained processing to 5 seconds using a timeout error, ensuring the code did not hang. The texts reduced from about 10,000 to 6,408 after this filtering.}. We then extracted a segment of text of 5,000 words, from the 5,000th to the 10,000th word. This was to ensure that we did not conduct RQA over the table of contents of the Gutenberg works. This facilitates analysis under limitations of memory, but it also ensures that genres are compared on approximately the same length of text. Before turning these word sequences into time series, we converted them into tibbles using the $\code{tidytext}$ tokenizer \cite{silge2016tidytext}.

\begin{figure}
\vspace*{0in}
  \centering
    \includegraphics[width=0.8\textwidth]{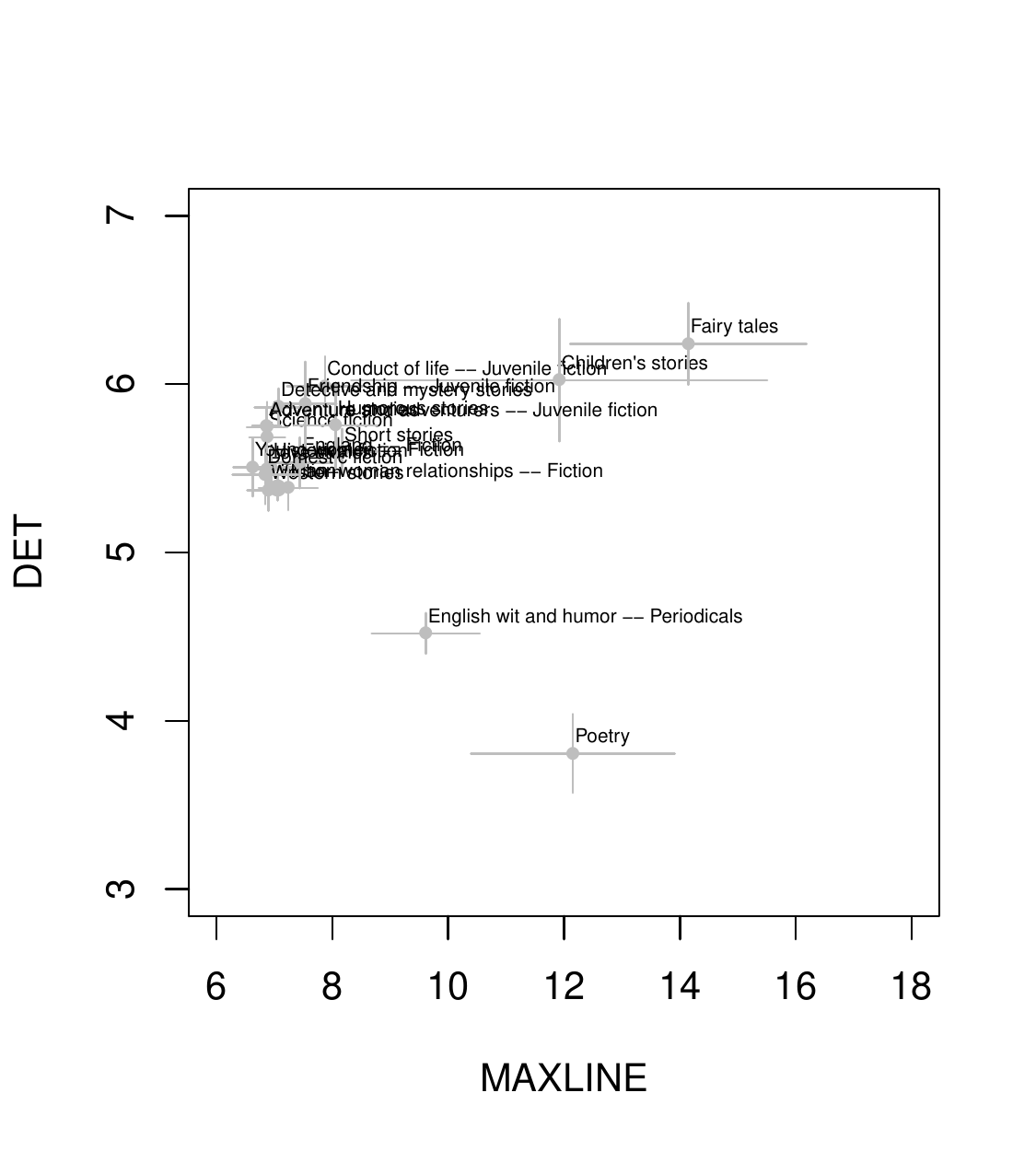} 
  \caption{An illustration of how genres extracted from the Gutenberg Project with \code{gutenbergr} cluster. Under the measures $MAXLINE$ and $DET$, we see that there is a core cluster of generic fiction, but an apparent tendency for children's literature and poetry to cluster differently, for example. Lines reflect 99\% confidence intervals using $N$ in Table \ref{tb:rqameasures}.}    
  \label{fg:gutenberg_max_det}
\end{figure}

\begin{figure}
\vspace*{0in}
  \centering
    \includegraphics[width=0.8\textwidth,trim={0 2cm 0 0},clip]{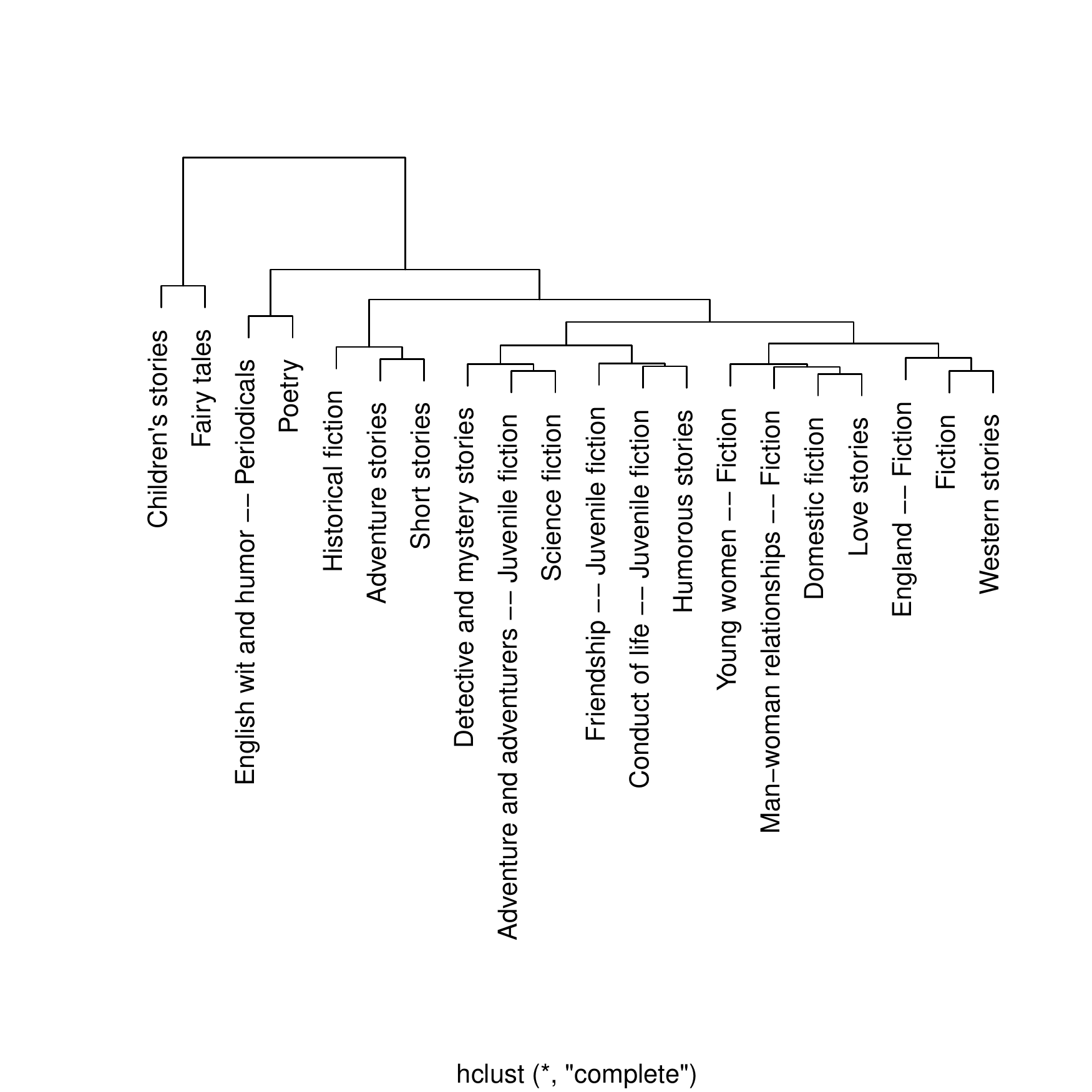} 
  \caption{Using measures shown in Table \ref{tb:rqameasures}, and standardizing them (by column), a dendrogram from the Euclidean distance matrix of these genres reveals their relationships. See text for details.}    
  \label{fg:gutengerg_dendrogram}
\end{figure}

Much like the example in the prior section, all texts were subjected to RQA. The results are shown in Table \ref{tb:rqameasures}. There is clear clustering when we look at the way that genres behave under different measures. An example is shown in Fig. \ref{fg:gutenberg_max_det}, showing that poetry, as might be expected, has very high $MAXLINE$, but surprisingly lower $DET$. When we use all these measures, and cluster them using a dendrogram \code{hclust} in \code{R}, we obtain what is shown in \ref{fg:gutengerg_dendrogram}. Poetry and fairy tales are in a separate cluster relative to the majority of the genres, with the general class of ``prose fiction'' seeming to reflect the mass of this distribution. However subtle features of the dendrogram suggest genres cluster together in intuitive ways. For example, ``juvenile fiction'' clusters in one leaf, and adventures and westerns are close in the tree. It is important to note that some of these results may be due to works falling under multiple categories. However, this only represents a small percentage of the overall distribution. Out of the approximately 6,400 texts, only 16\% are listed under two categories, and only 2\% under three.  

All measures are significantly accounted for by genre category. For each measures, $RR$, $DET$, etc., we build a separate regression model that predicts the value of those measures from 20 genres in the form of dichotomous variables. The $R^2$ values from these regression models are also shown at the bottom of Table \ref{tb:rqameasures}. All measures significantly vary in some manner around genre category, with the most variance accounted for in the $RR, DET, MEANLINE$ and $ENT$ measures, all over 10\% of variance accounted for by the 20 genre codes. 

In sum, the surface dynamics of a text, without any content analysis whatsoever, marks the classification of a text in these 20 genres in the Gutenberg Project. Follow-up analysis of this set should of course explore other factors that may be important. For example, some samples of poetry in Gutenberg reflect many poems--whereas here we assume the 5,000 words as a single text. A windowed analysis of the kind described above may also be a fruitful way of controlling for these greatly varying aspects of text length.

% Table created by stargazer v.5.2 by Marek Hlavac, Harvard University. E-mail: hlavac at fas.harvard.edu
% Date and time: Mon, Dec 18, 2017 - 15:26:42
\renewcommand{\arraystretch}{2.0}
\begin{table}[!htbp] \centering\scriptsize
  \caption{Means of RQA measures for the various genres extracted from the Gutenberg dataset (\code{gutenbergr}).} 
  \label{} 
\begin{tabular}{@{\extracolsep{6pt}} rcccccc} 
\\[-1.8ex]\hline 
\hline \\[-1.8ex] 
Genre & N & $RR$ & $DET$ & $MAXLINE$ & $MEANLINE$ & $ENT$ \\ 
\hline \\[-1.8ex] 
Adventure and adventurers -- Juvenile fiction & 244 & $0.97$ & $5.74$ & $6.87$ & $2.07$ & $0.24$ \\ 
Adventure stories & 584 & $1.03$ & $5.75$ & $6.86$ & $2.06$ & $0.23$ \\ 
Children's stories & 239 & $1.03$ & $6.02$ & $11.92$ & $2.10$ & $0.29$ \\ 
Conduct of life -- Juvenile fiction & 504 & $0.96$ & $5.99$ & $7.87$ & $2.07$ & $0.26$ \\ 
Detective and mystery stories & 460 & $0.96$ & $5.86$ & $7.07$ & $2.07$ & $0.24$ \\ 
Domestic fiction & 247 & $0.93$ & $5.46$ & $6.84$ & $2.06$ & $0.22$ \\ 
England -- Fiction & 272 & $0.95$ & $5.53$ & $7.44$ & $2.06$ & $0.22$ \\ 
English wit and humor -- Periodicals & 317 & $0.90$ & $4.52$ & $9.61$ & $2.06$ & $0.20$ \\ 
Fairy tales & 198 & $1.17$ & $6.24$ & $14.15$ & $2.13$ & $0.34$ \\ 
Fiction & 1115 & $0.97$ & $5.38$ & $7.06$ & $2.06$ & $0.22$ \\ 
Friendship -- Juvenile fiction & 252 & $0.93$ & $5.88$ & $7.53$ & $2.07$ & $0.26$ \\ 
Historical fiction & 426 & $1.07$ & $5.50$ & $6.94$ & $2.06$ & $0.20$ \\ 
Humorous stories & 201 & $0.96$ & $5.75$ & $8.05$ & $2.08$ & $0.26$ \\ 
Love stories & 520 & $0.94$ & $5.49$ & $6.87$ & $2.06$ & $0.22$ \\ 
Man-woman relationships -- Fiction & 307 & $0.92$ & $5.38$ & $7.24$ & $2.06$ & $0.21$ \\ 
Poetry & 231 & $0.97$ & $4.31$ & $10.94$ & $2.09$ & $0.24$ \\ 
Science fiction & 485 & $0.99$ & $5.68$ & $6.87$ & $2.07$ & $0.24$ \\ 
Short stories & 494 & $1.04$ & $5.60$ & $8.17$ & $2.07$ & $0.23$ \\ 
Western stories & 397 & $0.99$ & $5.37$ & $6.89$ & $2.06$ & $0.21$ \\ 
Young women -- Fiction & 239 & $0.90$ & $5.51$ & $6.62$ & $2.06$ & $0.22$ \\ 
\hline \\[-1.8ex]
$R^2$ of measure predicted &  & $0.11$ & $0.13$ & $0.02$ & $0.10$ & $0.11$ \\ 
\hline \\[-1.8ex] 
\end{tabular} 
\label{tb:rqameasures}
\end{table}

\section{Conclusions and future directions}

We have shown formal relationships between a common dynamic systems analysis framework, RQA, and traditional NLP, $n$-gram methods. These relationships are precise, and common measures in RQA can be recovered from NLP. These relations are not in each case simplistic. $DET$ and derived $ENTR, MAXLINE$, and $MEANLINE$ are based on maximally bounding sequences, reflecting an efficient encoding of the sequence of words. The RQA framework instead provides a basis for describing the patterns of \emph{dynamic revisitation} across a text: how often a text revisits words ($RR$), in what sequential structure ($DET$, etc.), with what orderliness ($ENT$), and so on.

These dynamic features of text lie at the ``surface'' -- we do not need to know the particular words of a text, but just the dynamic structure of their occurrence. Using these signatures, we can classify genre from the Gutenberg Project. 

Further use of RQA may be expanded into semantic levels of analysis, and multimodal analyses. In this final section, we briefly describe two ways we may extend these methods. First, we show that semantic vectors from neural network modeling can be used to build RQA analyses. Using the popular $\code{word2vec}$ framework \cite{mikolov2013distributed}, we show how an $RP$ from semantics can be constructed (see also \cite{angus2013making}). Following this, we briefly describe joint recurrence plots ($JRP$). These are constructed by the product of different $RP$s, and can be used to discern when different levels of analysis are co-occurring in a text, thus opening possibilities for multi-level dynamic analysis.

\subsection{Semantic recurrence: word2vec + RQA}

Neural network models have become a standard approach to word and document semantics over the past decade or so, perhaps most prominent among these models the $\code{word2vec}$ model \cite{mikolov2013distributed}. This model uses a predictive scheme in which $D$ hidden unit representations are trained. These $D$ units become a $D$-dimensional vector for each word seen the network. The relationship among these vectors can reveal fascinating semantic relations, all from simply training a neural model to predict word sequences.

It is easy to adapt representations of this sort for RQA, and we show a quick example here (code also included in GitHub repository). One of our texts may have a 5,000-word sequence, and after processing with the neural model ($\code{https://github.com/bmschmidt/wordVectors}$), we obtain a $\mathbf{M} = 5000 \times D$ matrix reflecting the text's ``trajectory'' through semantic space. A simple matrix norm of this sort can generate a distance metric: $|\mathbf{M_z}\mathbf{M_z^T}|$, where $z$ reflects scaled values for the $D$ dimensions. 

\begin{figure}
  \centering
    \includegraphics[width=0.5\textwidth]{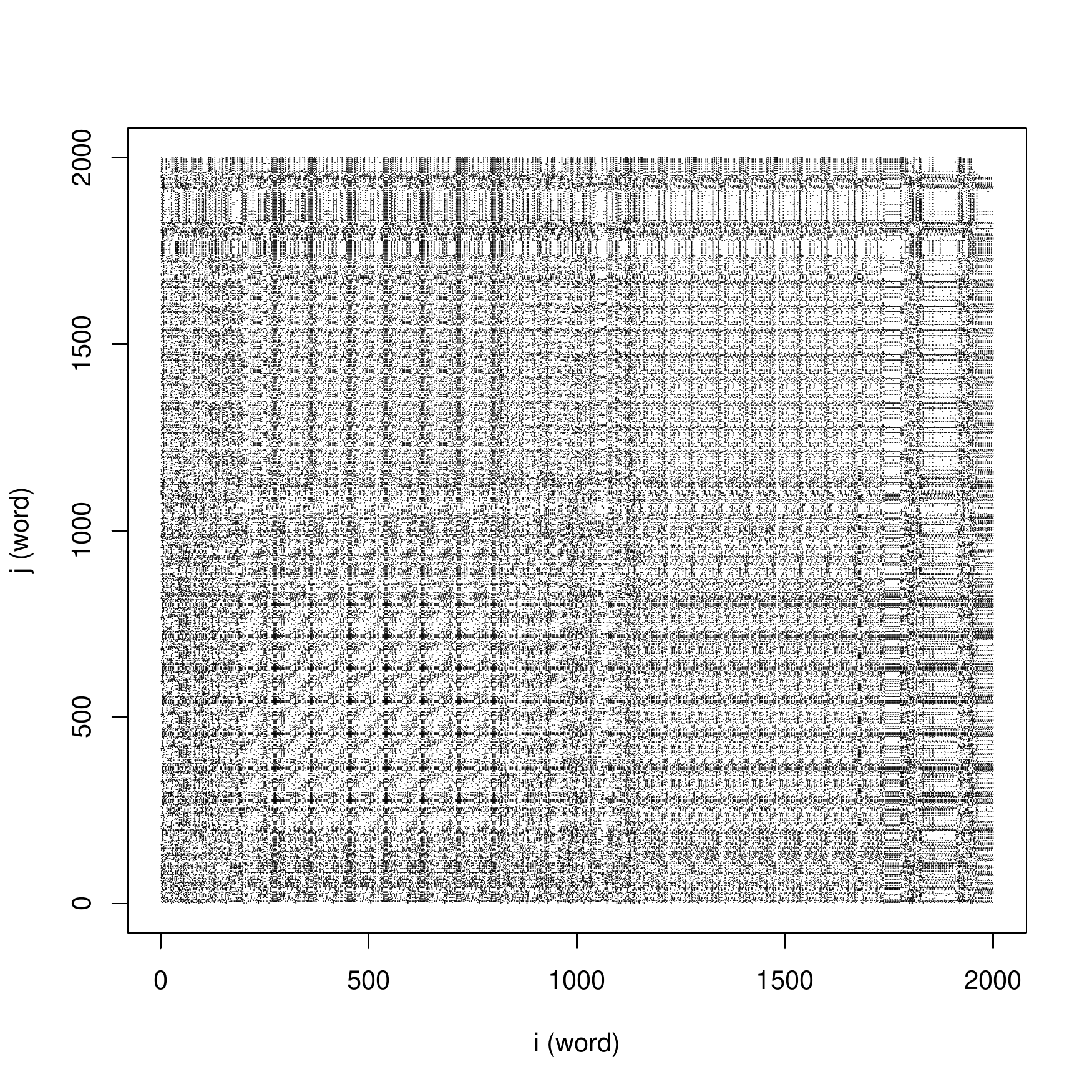}
  \caption{Sample \emph{semantic} $RP$ of a text segment of \emph{A Kindergarten Story} by Jane Hoxie using a $\code{word2vec}$ model of a subset of the Gutenberg books (see GitHub repository for model training).}    
  \label{semantic_rp}
\end{figure}

By setting a distance threshold for these semantic vectors, we obtain a ``semantic RP'' as shown in Fig. \ref{semantic_rp}. This uses the same 1906 text as that above, by Hoxie. Using the $\code{crqa}$ library's $\code{recpt}$ option on the main function call, we can compute all the regular RQA measures over this semantic representation. 

\subsection{Joint recurrence: the JRP}
 
A joint recurrence plot ($JRP$) is simply the product of multiple $RPs$: $JRP = \prod_i RP_i$. The logic of a $JRP$ is that levels of analysis (such as, say, semantics and syntax) will show interesting levels of co-occurrence. The product of the $RP$ will return a sparser plot containing points $(i,j)$ where multiple levels or ``modalities'' overlapped in occurrence. A full example using the $\code{crqa}$ library is developed here: $\code{https://github.com/racdale/jrp-example}$.

% show measures

% RR is this in n-gram analysis

% DET is this in n-gram analysis

% max line and mean line are this

% entropy is this

% should we get TWO texts and show expected differences?

\bibliographystyle{splncs_srt}
\bibliography{bib}
\end{document}